\documentclass{egpubl}
\usepackage{pg2024}

\SpecialIssuePaper         

\CGFStandardLicense

\usepackage[T1]{fontenc}
\usepackage{dfadobe}  

\usepackage{multirow}
\usepackage{array}  
\newcolumntype{C}[1]{>{\centering\arraybackslash}p{#1cm}}
\usepackage{amssymb} 
\usepackage{amsmath} 
\usepackage{booktabs} 

\usepackage{cite}  
\BibtexOrBiblatex
\electronicVersion
\PrintedOrElectronic
\ifpdf \usepackage[pdftex]{graphicx} \pdfcompresslevel=9
\else \usepackage[dvips]{graphicx} \fi

\usepackage{egweblnk}

\title[P-Hologen: An End-to-End Generative Framework for Phase-Only Holograms]%
      {P-Hologen: An End-to-End Generative Framework for Phase-Only Holograms}

\author[JH Park et al.]
{\parbox{\textwidth}{\centering 
JooHyun Park$^{1}$\orcid{0009-0009-5891-7405}, 
YuJin Jeon$^{2}$\orcid{0009-0001-6214-7302}, 
HuiYong Kim$^{1}$\orcid{0000-0001-7308-133X}, 
SeungHwan Baek$^{2}$\orcid{0000-0002-2784-4241},
and HyeongYeop Kang$^{3*}$\orcid{0000-0001-5292-4342
} 
        }
        \\
{\parbox{\textwidth}{\centering $^1$ Kyung Hee University, South Korea\\
         $^2$ POSTECH, South Korea \\
         $^{3*}$ Korea University, South Korea
       }
}
}

\begin{document}


\maketitle
\begin{abstract}
Holography stands at the forefront of visual technology, offering immersive, three-dimensional visualizations through the manipulation of light wave amplitude and phase. 
Although generative models have been extensively explored in the image domain, their application to holograms remains relatively underexplored due to the inherent complexity of phase learning.
Exploiting generative models for holograms offers exciting opportunities for advancing innovation and creativity, such as semantic-aware hologram generation and editing.
Currently, the most viable approach for utilizing generative models in the hologram domain involves integrating an image-based generative model with an image-to-hologram conversion model, which comes at the cost of increased computational complexity and inefficiency.
To tackle this problem, we introduce P-Hologen, the first end-to-end generative framework designed for phase-only holograms (POHs). 
P-Hologen employs vector quantized variational autoencoders to capture the complex distributions of POHs. 
It also integrates the angular spectrum method into the training process, constructing latent spaces for complex phase data using strategies from the image processing domain. 
Extensive experiments demonstrate that P-Hologen achieves superior quality and computational efficiency compared to the existing methods. 
Furthermore, our model generates high-quality unseen, diverse holographic content from its learned latent space without requiring pre-existing images. 
Our work paves the way for new applications and methodologies in holographic content creation, opening a new era in the exploration of generative holographic content.
The code for our paper is publicly available on https://github.com/james0223/P-Hologen.


\end{abstract}  
\section{Introduction}
\label{sec:intro}

Holography is an emerging technology that unveils a realm where the conventional boundaries of image capturing and representation are transcended. 
Traditional photography is confined to recording the intensity of light, making it lose three-dimensionality and present a two-dimensional portrayal of the world. 
In contrast, holography is able to encapsulate not just the amplitude, but also the phase values of light waves, thereby preserving the spatial information intrinsic to every scene.

Unfortunately, the pace of evolution in the holography domain is slower than that in the image domain due to the technical complexity, expensive equipment, computational intensity, and material limitations~\cite{shimobaba2015review, blanche2021holography}.
To overcome this, many researchers begin with simpler forms of holography such as 2D phase-only holograms (POHs).
POHs simplify the data handling process by keeping the amplitude of the wavefront constant and only considering the phase variants information to encode the information of the scene.
Therefore, researchers conduct experiments and test their hypothesis with the 2D POHs to obtain a foundational understanding of light's behavior and the principles of interference and diffraction~\cite{peng2020neural, liu2021deep}. The gained understanding serves as a prerequisite for exploring the more complex domain of 3D holography. 

These days, the research topic in the image domain has begun to include generative models for 2D images. 
Such models have transformative potential in synthesizing realistic and high-quality visuals, augmenting datasets, and enhancing various applications such as superresolution and text-driven modifications~\cite{singh2023high, gao2023implicit}. 
However, due to its complexity, only a few attempts have been made to translate this success to the realm of holography~\cite{liu2021deep}.
In terms of phase data learning, generative models have to process intricate patterns and dependencies between different phases and phase wraps that can cause abrupt changes and discontinuities in the phase data. Furthermore, developing appropriate metrics and loss functions that consider the unique characteristics of phase information becomes a challenge.
These contrast with the simple, pixel-based intensity values commonly encountered in 2D images, making it challenging to train generative models for holography.

As for now, there exist two approaches to exploit the advanced features of generative models for novel POH generation: training a generative model with Spatial Spectrum of Hologram Modulators (SSHMs) introduced by Liu \textit{et al.}~\cite{liu2021deep}, and the integration of an image generative model~\cite{singh2023high, tumanyan2023plug} with an image-to-hologram conversion model~\cite{peng2020neural, wang2020phase, wu2021high, zhang2021phasegan}.

The first approach has demonstrated effective performance in generating in-between POH frames for a sequence of POH animation frames~\cite{liu2021deep}. However, this method is effective only in relatively simple data scenarios, as will be presented in~\autoref{subsec:comparative}.
The second strategy is capable of producing high-quality holograms, but the integration of two distinct models requires more memory space for model parameters, a complex task involved in training two distinct models, and extended generation times from the necessity of passing data through both models sequentially. 

\begin{figure}[t]
    \centering
    \includegraphics[width=\columnwidth]{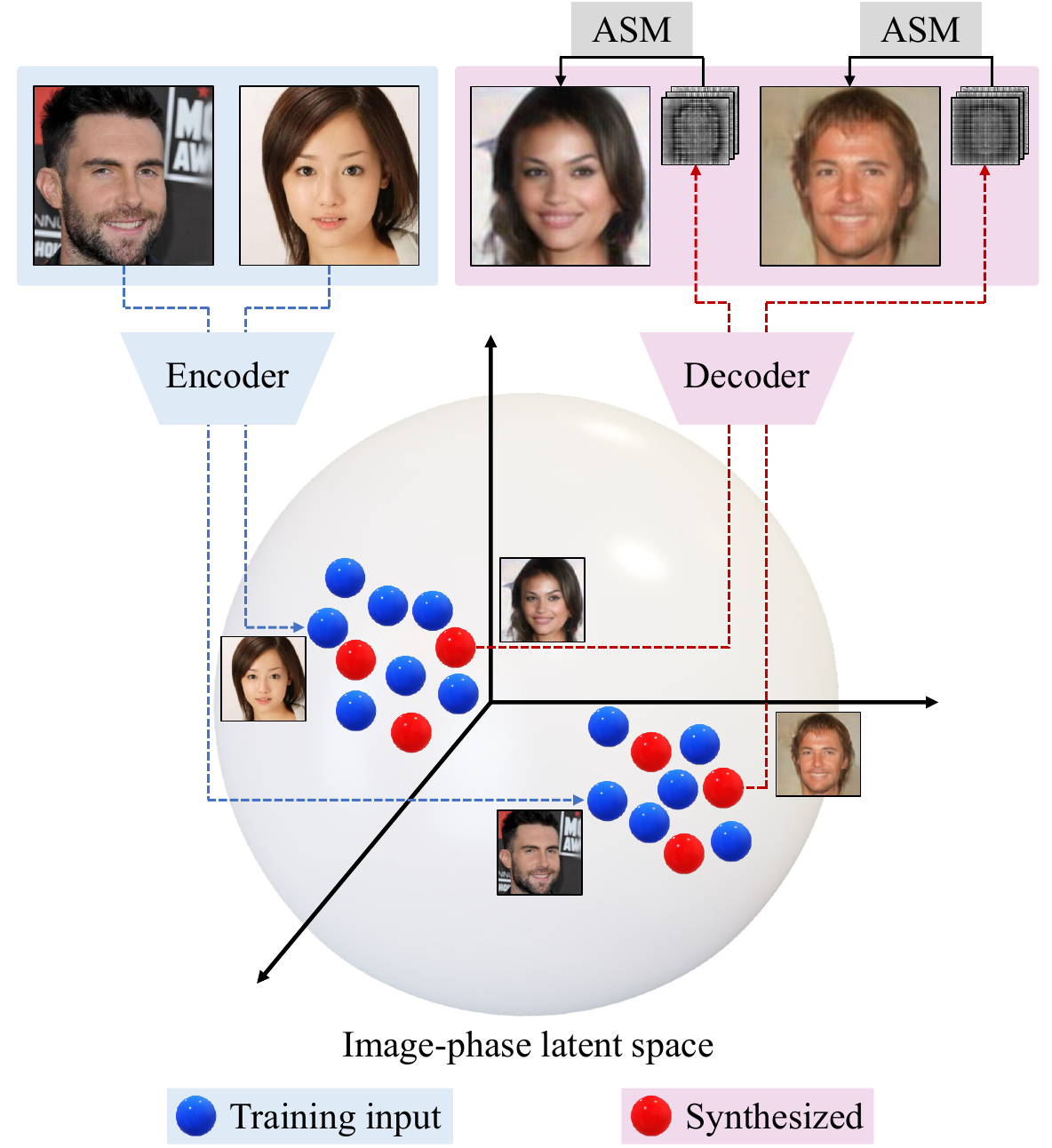}
    \caption{During the training phase, P-Hologen learns image-phase latent representations from an image dataset. During inference, POHs can be generated by sampling the image-phase latent space. These sampled POHs are then propagated using the ASM to reconstruct previously unseen images.}
    \label{fig:teaser}
\end{figure}

In this paper, we introduce P-Hologen, a novel end-to-end generative model architecture for 2D POH generation. 
Inspired by Holonet~\cite{peng2020neural}, we demonstrate that leveraging images as input and computing the reconstruction loss in the image domain enables the model to learn the data distribution and sample novel POHs. 
Specifically, we apply the Angular Spectrum Method (ASM) on the generated POH to reconstruct an image, which is then compared with the input image to compute the reconstruction loss.

In the image domain, Vector-Quantized Variational Autoencoders (VQ-VAEs)~\cite{van2017neural, razavi2019generating} are recognized for producing improved generative results compared to conventional Variational Autoencoders (VAEs). 
This advantage is attributed to VQ-VAEs constructing a discrete latent space, which more effectively represents the discrete features inherent in most types of data~\cite{van2017neural}. 
Similarly, we found that VQ-VAEs produced superior results compared to VAEs for POHs through the experiments, as presented in~\autoref{subsec:vq_analysis}. 
Therefore, we implement the P-Hologen based on the VQ-VAE framework.
As illustrated in~\autoref{fig:teaser}, the VQ-VAE model constructs discrete image-phase latent representations of POHs from image inputs, facilitating the generation of novel POHs.

In summary, the contributions of our paper are as follows:
\begin{itemize}
    \item We introduce P-Hologen, the first end-to-end generative model architecture for unseen POH generation, trained with novel learning strategies and loss functions.
    \item We propose a streamlined hologram generative approach that requires less memory and generation time, yielding improved reconstructed quality compared to existing methods that combine separate image generative and image-to-hologram conversion models.
    \item We conduct an extensive study demonstrating the potential and superiority of P-Hologen over existing methods.
    \item We explore the potential of a generative approach within the POH domain, anticipating its extension to generative models for POHs, capable of accommodating variations in parameters such as propagation distance and pixel pitch.
\end{itemize}

\begin{figure*}[t]
    \centering
    \includegraphics[width=\textwidth]{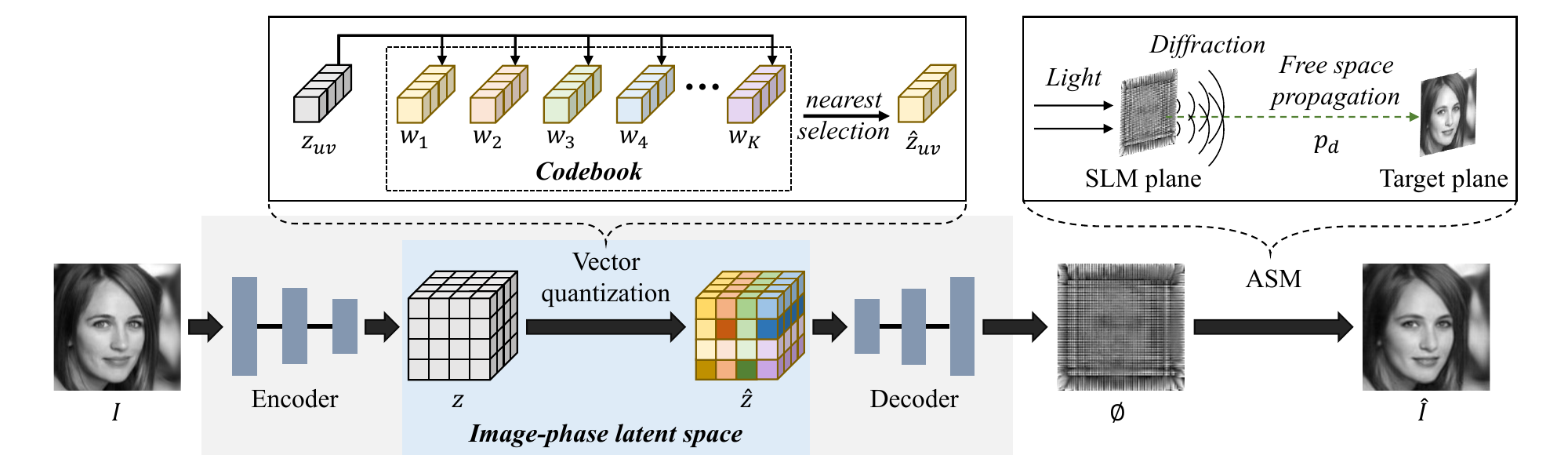}
    \caption{Overall architecture of P-Hologen}
    \label{fig:overview}
\end{figure*}

\section{Related works}
\label{sec:rel_work}

Holograms are fringe patterns that encapsulate the intricate interplay of light waves reflected from a scene. 
The hologram itself is a two-dimensional surface that contains information about the intensity and phase distribution of light waves, but the image it produces appears three-dimensional and can be viewed from different angles.

Holograms can be produced through two primary methods: optical generation and computational generation, with the latter referred to as Computer-Generated Holography (CGH). 
In recent years, CGH has gained increasing prominence over optical methods, primarily due to the stringent environmental conditions and precise control required for the generation of optical holograms.
Additionally, CGH can create holograms of synthetic scenes, offering much richer holographic content.

In the idealized conception of CGH, both amplitude and phase play pivotal roles, collaboratively contributing to the multi-dimensional portrayal of scenes with enhanced depth and perspective. 
However, due to the technological complexities of spatial light modulators (SLMs) in processing complex holograms that contain both amplitude and phase, a substantial corpus of holographic research often focuses on either amplitude-only holograms or POHs~\cite{jiao2018compression}.

Among these, POHs have gained more research attention. This inclination is attributed to the inherent capacity of POHs to yield reconstructions marked by enhanced brightness and clarity~\cite{pang2017non, tsang2021optimal}. 
The process of generating POH is intricately connected to phase retrieval, a technique that focuses on deducing the phase of input data from the magnitude of its Fourier transform~\cite{shechtman2015phase}.
Phase retrieval is a complex, non-linear, and ill-posed problem. There are inherent ambiguities and constraints that make the retrieval of phase information challenging. 
For this reason, a plethora of algorithms exist in the realm of phase retrieval, each tailored to address specific aspects of these challenges.

Traditionally, algorithmic approaches have long been studied. They are inherently systematic and are grounded in well-defined mathematical principles. Algorithmic approaches can indeed be primarily classified into iterative and non-iterative algorithms. Iterative algorithms for phase retrieval involve a repetitive process of refinement to achieve an accurate estimation of the phase~\cite{gerchberg1972practical, bengtsson1994kinoform, yoshikawa1995quantized, dresel1996design, jabbour2008vectorial, peng2017mix}.
On the other hand, non-iterative algorithms are characterized by their ability to provide rapid phase estimations, resulting in significantly reduced computation times~\cite{lohmann1967binary, tsang2013novel, tsang2014generation, mendoza2014encoding}.  
However, they are often hampered by issues related to computational intensity, convergence challenges, sensitivity to noise, and the expertise required for implementation.

Recently, the advent and proliferation of deep learning techniques have indeed instigated notable advancements in the field of POH generation. 
Deep learning facilitates the rapid generation of high-quality POHs by learning intricate patterns of light wave interactions within vast datasets, leading to phase hologram generation that is not only accurate but also adaptable across diverse scenarios and conditions~\cite{chakravarthula2019wirtinger, peng2020neural, hossein2020deepcgh, chakravarthula2020learned, wang2020phase, kang2021deep, choi2021neural, wu2021high, zhang2021phasegan, shi2021towards, shi2022end, yang2022diffraction, kavakli2022learned, ban2024nhvc}.

The recent integration of deep learning in POH generation has seen significant strides, yet the incorporation of generative models has not been as pronounced. 
Generative models, characterized by their sophisticated latent space representations, offer advanced techniques for data manipulation~\cite{singh2023high, tumanyan2023plug}. The integration of these models into POH generation holds the potential to expand the horizon of possibilities in holography, paving the way for innovative applications and methodologies.

One relevant study is Liu \textit{et al.}~\cite{liu2021deep} that introduces a generative approach for hologram generation named channeled variational encoder. 
This involves an assumption where the target hologram is modeled as the product of the original hologram and a specific transformation function. 
While this method circumvents the complexities associated with mapping intricate phase data into a latent space, it inherently reveals a limitation to generating more complex scenes that cannot be derived simply by applying a transformation function. 
Our model markedly deviates from Liu's paradigm, embodying an enhanced capability to directly synthesize unseen POHs from a learned latent space.

\section{Method}
\label{sec:method}

The overall pipeline of our model, P-Hologen, is illustrated in~\autoref{fig:overview}.
It primarily consists of an encoder and a decoder parts.
During the training phase, the encoder's role is to convert an input 2D image into a corresponding representation within the latent space. 
Subsequently, the decoder is tasked with the generation of a POH from this latent representation.
The objective function regarding reconstruction is computed in the image domain, which aims to minimize the discrepancy between the original input image and the image reconstructed from the POH.
The architecture of our model is based on the work by Esser \textit{et al.}~\cite{esser2021taming}, which utilizes a discrete latent space~\cite{van2017neural} to effectively encapsulate the distinctive features of the data.
This approach enables a condensed yet richly detailed encoding of the input, ensuring that the essential characteristics are efficiently represented within the image-phase latent space.
The details on model configurations are illustrated in the supplementary material.

\subsection{Encoding}
\label{subsec:encoding}
The encoder processes a 2D image $I \in \mathbb{R}^{N \times N \times C}$, where $N$ denotes the image's width and height and $C$ denotes the number of channels.
The encoding encompasses the convolution of input features and their subsequent downsampling to yield the latent vector $z \in \mathbb{R}^{n \times n \times H}$, where $n$ denotes its width and height, and $H$ specifies its number of channels.

To construct the latent space for complex phase data, we decide to use 2D images as the training input rather than POHs, with the output being a POH. This decision brings two primary advantages. 

Firstly, it enables us to leverage established training strategies from image processing domains, avoiding the need to develop new, complex strategies for phase data learning. Liu \textit{et al.}~\cite{liu2021deep} highlighted the significant challenges of directly learning from phase data, noting that its inherent complexity and sensitivity to errors can severely hinder the learning process, often resulting in considerable loss in reconstructed outputs from minor inaccuracies.

Secondly, it permits the use of high-quality image datasets for training instead of POH datasets. This circumvents the substantial training barrier posed by the scarcity of high-quality POH datasets~\cite{zeng2021deep}. Although conversion models exist that transform standard image datasets into POH datasets~\cite{peng2020neural, zhang2021phasegan}, the quality of the resulting holographic data is often inferior to the original images, and the conversion process can be time-consuming. Maintaining high-quality training data is crucial, as any reduction in quality can adversely affect the performance of the trained model.

\subsection{Vector quantization}
\label{subsec:vec_quantize}
The vector quantization process is a crucial step in VQ-VAEs, distinguishing them from standard VAEs by quantizing a continuous latent vector $z$ into a discrete counterpart $\hat{z}$. 
This is achieved using a codebook consisting of $K$ representative vectors.
For each pixel $z_{uv}$ in the continuous latent space, where $z_{uv} \in \mathbb{R}^{ 1 \times 1 \times H }$ and  $u, v \in [1, n]$, we measure its similarity to each codebook vector $w_{k} \in \mathbb{R}^{ 1 \times 1 \times H }$, where $k\in K$. 
The similarity is assessed by computing the squared Euclidean distance.
Then, the codebook vector $w_{k}$ that is most similar to $z_{uv}$ is then chosen as the quantized counterpart $\hat{z}_{uv} \in \mathbb{R}^{ 1 \times 1 \times H }$. Mathematically, this selection process is defined as follows:
\begin{equation}
    \hat{z}_{uv} = \underset{w_k}{\mathrm{argmin}} \ \| z_{uv} - w_k \|^2
    \label{eq:quantization}
\end{equation}

The quantization process is applied across all pixels, transforming the continuous latent vector $z$ into its discrete equivalent $\hat{z}$, thereby facilitating the generation of a discrete latent representation of the input data. 

\subsection{Decoding and reconstruction}
\label{subsec:decoding_recon}
During the decoding stage, the quantized vector $\hat{z}$ is processed through a series of convolutional layers and upsampling steps to construct a POH $\phi \in \mathbb{R}^{ N \times N \times C }$.

The produced POHs encapsulate phase information in the form of intricate fringe patterns, which do not present a recognizable depiction of the original scene. To convert the patterns back into a visual representation of the scene, a numerical reconstruction process is essential.
This typically involves simulating the wavefront propagation, effectively retracing the path of light as captured in the hologram.

For wavefront propagation, our model utilizes the ASM~\cite{goodman2005introduction, matsushima2009band}, a technique renowned for accurately simulating light wave propagation. The application of ASM to $\phi$ is defined by the following equations:

\begin{align}
    \begin{split}
        \hat{\phi_{c}}(p_{d}) = \iint F\left(e^{i\phi_{c}}\right)H(f_x, f_y,p_{d}) e^{i2\pi(f_x x + f_y y)}df_x df_y ,
        \\\\
        H(f_x, f_y, p_{d}) = 
        \begin{cases}
        e^{i\frac{2\pi}{\lambda_{c}} \sqrt{1 - \lambda_{c}^2 (f_x^2 + f_y^2)}p_{d}}, & \text{if} \sqrt{f_x^2 + f_y^2} < \frac{1}{\lambda_{c}}, \\
        0 & \text{otherwise}
        \end{cases}
    \end{split}
    \label{eq:asm}
\end{align}
where $p_{d}$ denotes the propagation distance, ${\phi}_{c}$ denotes the $c \in C$ channel of ${\phi}$, $\hat{\phi}_{c}$ denotes the propagated scalar field of ${\phi}_{c}$ at distance $p_{d}$, $F$ denotes the Fourier Transform, $\lambda_{c}$ denotes the wavelength corresponding to channel $c$,  $x \in [1, N]$ and $y \in [1, N]$ denote the spatial coordinates, and $f_{x}$ and $f_{y}$ denote the spatial frequency components of ${\phi}_{c}$ in the $x$ and $y$ directions.
These equations describe the wavefront's transformation as it propagates from the spatial light modulator (SLM) plane, through free space, undergoing diffraction, and finally reaching the target plane, where the holographic image is formed.
It is important to note that ASM is applied separately to each color channel, as the wavelength $\lambda_{c}$ varies for different colors in the light spectrum. 

The propagated scalar field $\hat{\phi}$ contains both amplitude and phase information.
Here, the phase information does not directly contribute to the perceived image but rather to the way light interferes to form that image. In contrast, the amplitude pattern is what needs to be visualized to represent the reconstructed image from a hologram.
Therefore, it is essential to extract the amplitude patterns of $\hat{\phi}$.
The amplitude extraction is defined by the following equations: 
\begin{equation}
    \hat{I} = s \cdot |\hat{\phi}| 
    \label{eq:amplitude}
\end{equation}
where $\hat{I}$ denotes the reconstructed image and $s$ denotes a hyperparameter representing a variable for considering the gap between the output values of the wave propagation operator (ASM) and the desired target~\cite{peng2020neural}.

\subsection{Objective function}
\label{subsec:obj_funcs}
The P-Hologen model's objective function comprises three components: reconstruction loss ($L_{\mathrm{recon}}$), codebook loss ($L_{\mathrm{codebook}}$), and commitment loss ($L_{\mathrm{commit}}$). 

The reconstruction loss, $L_{\mathrm{recon}}$, quantifies the fidelity of the reconstructed output $\hat{I}$ compared to the original input $I$. 
It consists of a Mean Squared Error (MSE) loss $L_{\mathrm{mse}}$ and a perceptual loss $L_{\mathrm{per}}$~\cite{johnson2016perceptual}.
Our perceptual loss is determined by inputting both the original and reconstructed images into a pre-trained VGG16 network and comparing their feature maps extracted from ReLU 2-2 and ReLU 4-3.
This design ensures a comprehensive comparison by capturing both low-level features such as edges and textures from early layers, and high-level semantic features from deeper layers~\cite{rad2019srobb}. 

Our reconstruction loss $L_{\mathrm{recon}}$ is computed as a weighted sum of the MSE loss $L_{\mathrm{mse}}$ and the perceptual loss $L_{\mathrm{per}}$:
\begin{equation}
    L_{\mathrm{recon}} = \alpha \cdot L_{\mathrm{mse}}(I, \hat{I}) + \beta \cdot L_{\mathrm{per}}(I, \hat{I})
    \label{eq:recon_loss}
\end{equation}
where $\alpha=0.9$ and $\beta=0.1$ are empirically determined weights that mitigate the noise artifacts within the generated POHs. 
Experimental details are provided in~\autoref{subsec:recon_loss}. 

The codebook loss, $L_{\mathrm{codebook}}$, updates the codebook vectors to better approximate the encoder's outputs, effectively capturing the data distribution:
\begin{equation}
    L_{\mathrm{codebook}} = L_{\mathrm{mse}}(sg[z] - \hat{z})
    \label{eq:codebook_loss}
\end{equation}
where $sg$ denotes the stop-gradient operation that prevents gradient information from flowing back. 

The commitment loss, $L_{\mathrm{commit}}$, optimizes the encoder to produce latent vectors closely aligned with the codebook vectors.
This loss component restricts the latent space from becoming excessively expansive and ensures its conformity to a certain embedding:
\begin{equation}
    L_{\mathrm{commit}} = L_{\mathrm{mse}}(z - sg[\hat{z}])
    \label{eq:commit_loss}
\end{equation}

The aggregate objective function of P-Hologen is therefore the summation of these individual losses:
\begin{equation}
    L_{\mathrm{total}} = L_{\mathrm{recon}} + L_{\mathrm{codebook}} + L_{\mathrm{commit}}
    \label{eq:total_loss}
\end{equation}

\subsection{Sampling}
\label{subsec:sampling}
In a standard VAE, the latent space is modeled as a continuous Gaussian distribution, allowing straightforward sampling.
Novel data generation can be achieved by sampling a vector from the Gaussian distribution and passing it to the decoder for reconstruction.

In contrast, VQ-VAEs exploit a discrete latent space where the decoder input is represented by a combination of the codebook vectors.
Direct sampling from this discrete space is complex, necessitating an additional model to sample appropriate sequences of codebook vectors for generating new samples.
In our implementation, we select PixelSnail~\cite{chen2018pixelsnail} as the sampling model.

The PixelSnail model is trained on the discrete latent vectors $\hat{z}$ obtained from the pretrained VQ-VAE architecture. 
Specifically, the model utilizes the indices of the latent vectors $\hat{z}_{\mathrm{indices}}$ in the codebook, rather than the raw $\hat{z}$ values for computational efficiency.
During training, the $\hat{z}_{\mathrm{indices}}$ vectors are fed into the model to output a set of logits $\hat{z}_{\mathrm{predicted}}$ corresponding to the probability distribution over all possible quantized indices for each position in the $\hat{z}_{\mathrm{indices}}$ grid.
The predicted logits are compared with the ground truth $\hat{z}_{\mathrm{indices}}$ using the cross-entropy loss, allowing the model to learn the distribution of the latent vectors.
After training, the latent vectors sampled from the PixelSnail model are fed to the decoder of the pretrained VQ-VAE to produce novel data instances resembling the training data.
The details on PixelSnail configurations are illustrated in the supplementary material.

\section{Experiments}
\label{sec:evaluations}
The experiments were conducted leveraging datasets of MNIST \cite{lecun2010mnist}, CelebA-HQ~\cite{karras2017progressive} and cat images from Animal Faces HQ~\cite{choi2020stargan}, which we refer to as AF-C.
The MNIST dataset was resized to a resolution of $64\times64$ and was assigned a hologram configuration of a pixel pitch of 6.4 $\mu m$ and a propagation distance $p_{d}$ of 15 \textit{mm}.
The CelebA-HQ and AF-C datasets were resized to a resolution of $128\times128$ and were assigned a hologram configuration of a pixel pitch of 6.4 $\mu m$, and the propagation distance $p_{d}$ of 21.5 \textit{mm}.
The hyperparameter variable $s$ for amplitude extraction was set to $0.95$, following the implementation of Holonet~\cite{peng2020neural}.
The value of $K$ for the VQ-VAE architecture was fixed to 512 in accordance with the original paper, as it demonstrated the best performance in our internal tests.
The Adam optimizer was employed with a learning rate of $0.0002$, and the model was trained on each dataset for $100$ epochs using two NVIDIA Geforce RTX 3090 GPUs.
Further details can be found in the supplementary materials.

The experiment results were assessed using Peak Signal-to-Noise Ratio (PSNR), Structural Similarity Index Measure (SSIM), Fréchet Inception Distance (FID), and Learned Perceptual Image Patch Similarity (LPIPS) metrics.

\subsection{VQ-VAE for POH generation}
\label{subsec:vq_analysis}
 
The VQ-VAE architecture, introduced by Van \textit{et al.}~\cite{van2017neural}, is based on the hypothesis that a discrete latent space can more effectively represent the discrete features inherent in most data types. Building on this advantage, we leveraged VQ-VAE to learn the discrete latent space for POH generation.

However, unlike traditional images, which predominantly consist of consistent, low-frequency components, POHs contain a higher proportion of random, high-frequency areas. Although our model exploits images as inputs, the generated outputs are POHs. Consequently, the complex structure of the output phase data may pose a challenge for VQ-VAEs in developing a well-distributed latent space.

In this context, it is essential to explore whether the discrete latent space of VQ-VAEs can perform as effectively for POH generation as it does for image generation. To this end, we implemented VAE and VQ-VAE architectures based on the work of Esser \textit{et al.}~\cite{esser2021taming} and trained them for both image-to-image and image-to-POH generation tasks. The VAE model was trained using the conventional method, which constrains its latent space to a Gaussian distribution through the KL-divergence loss, while also incorporating our reconstruction loss presented in~\autoref{subsec:recon_loss}.

Model assessment was conducted through latent space visualization and quantitative evaluation using PSNR, SSIM, and FID metrics. 
To examine the latent spaces across different classes within the training dataset, we trained the models on the MNIST dataset, which consists of ten discrete classes of digit images. 
PSNR and SSIM scores were computed and averaged across all images in the validation set, while FID scores were obtained by comparing 100 generated samples to a randomly selected subset of 100 images from the validation set.
\begin{figure}[t]
    \centering
    \includegraphics[width=\columnwidth]{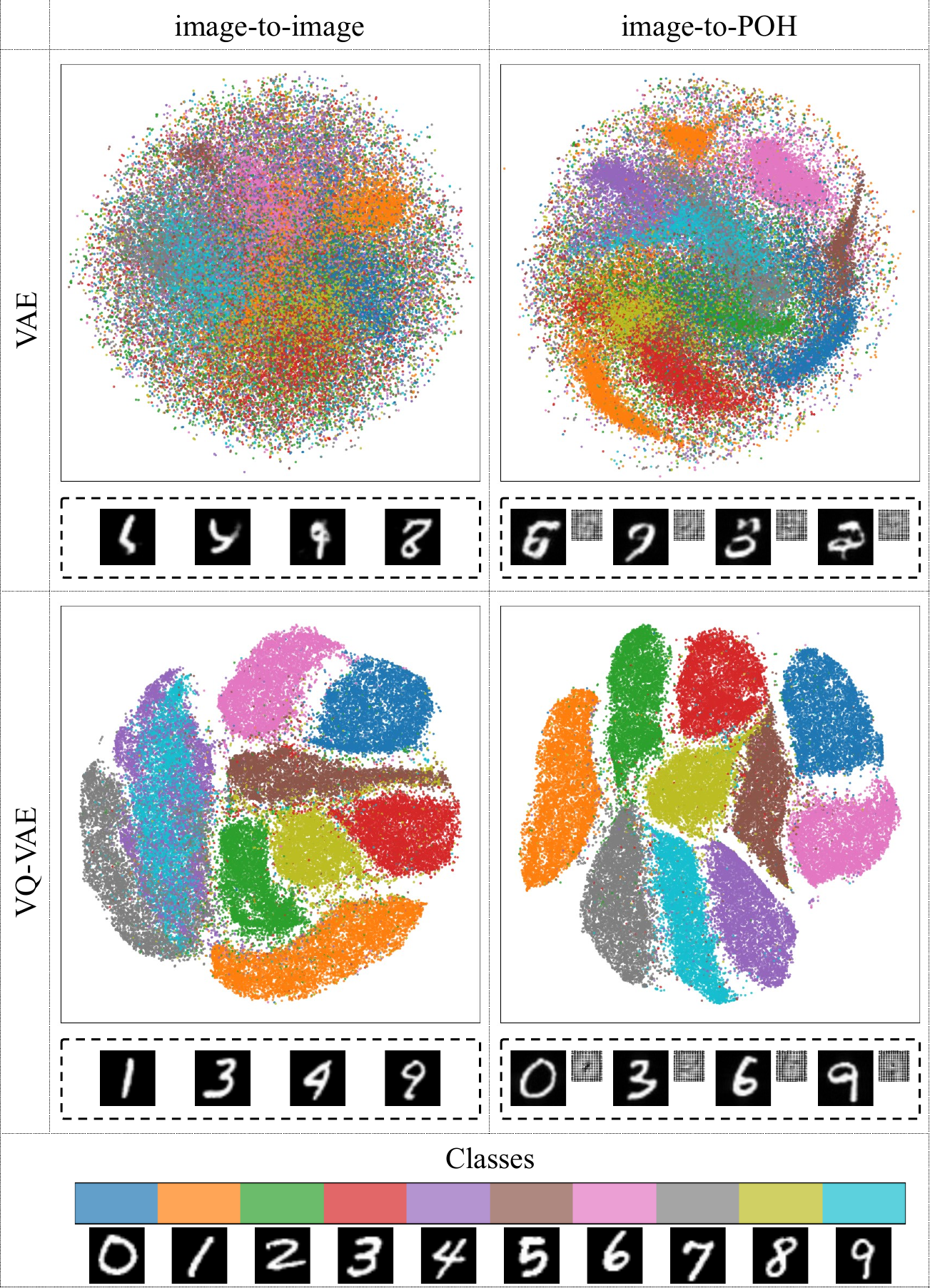}
    \caption{Visualization of the latent spaces trained on four different approaches using the MNIST dataset. The rows represent the architecture, and the columns represent the task. The images inside the dashed box represent the novel data instances sampled from each latent space.}
    \label{fig:latents_samples}
\end{figure}

\begin{table}[t]
    \centering
    \begin{tabular}{C{2.2}C{1.2}C{1.6}C{1.6}}
        \toprule
        Task         & Metric   & VAE & \textbf{VQ-VAE} \\ \midrule
        \multirow{3}{*}{image-to-image} 
            & PSNR($\uparrow$)            & 24.74   & 48.03    \\ 
            & SSIM($\uparrow$)            & \! \! \! \! \! \! 0.88     & \! \! \! \! \! \!  0.99     \\ 
            & FID($\downarrow$)            & 95.60     & 59.50     \\ \midrule
        \multirow{3}{*}{\textbf{image-to-POH}} 
            & PSNR($\uparrow$)            & 28.57   & \textbf{31.86}    \\ 
            & SSIM($\uparrow$)            & \! \! \! \! \! \! 0.68     & \! \! \! \! \! \! \textbf{0.71}     \\ 
            & FID($\downarrow$)            & 162.79 \,     & \textbf{102.13} \,     \\ \bottomrule
    \end{tabular}%
    \caption{Reconstruction accuracy and generation quality of VAE and VQ-VAE architectures trained on the MNIST dataset for image-to-image generation and image-to-POH generation tasks.}
    \label{tab:vae_vqvae}
\end{table}

The latent space of each approach, visualized using t-SNE~\cite{van2008visualizing}, is presented in~\autoref{fig:latents_samples}, accompanied by samples generated from the corresponding latent spaces. 
The visualizations indicate that the discrete latent space of VQ-VAEs, which exhibited clear class separation in the image-to-image generation task, was similarly well-distributed in the image-to-POH generation task, outperforming VAEs in both tasks. 
The quantitative evaluations presented in~\autoref{tab:vae_vqvae} support this observation, showing that VQ-VAEs outperform VAEs in both reconstruction accuracy and generated sample quality.

These results led us to conclude that VQ-VAEs are more effective than VAEs in learning features for image-to-POH generation, mirroring their superior performance in image-to-image generation. We hypothesize that although POHs exhibit high feature randomness within the phase domain, incorporating ASM propagation into the learning process helps structure the latent space based on the features present in their image domain reconstructions.

Additionally, while the latent space of the image-to-image VQ-VAE struggles to distinguish between the digits 4 and 9, the latent space of the image-to-POH VQ-VAE provides a relatively clear separation between these digits. This observation suggests that integrating the propagation simulation with the VQ-VAE architecture enhances the understanding of features in the training dataset, further justifying our selection of VQ-VAEs for POH generation.

\begin{figure}[t]
    \centering
    \includegraphics[width=\columnwidth]{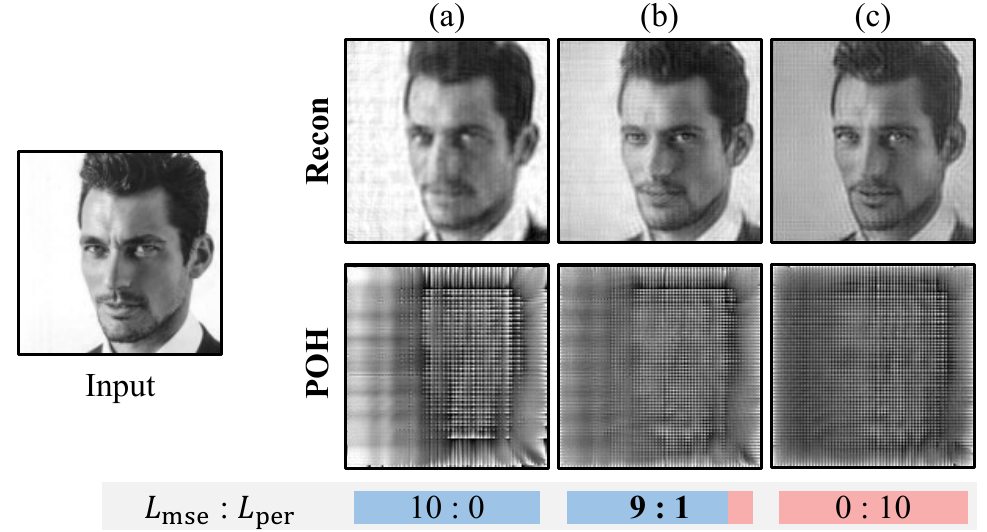}
    \caption{Illustration of POH reconstructions using different loss functions: (a) $L_{\mathrm{mse}}$ only; (b) a 9:1 ratio of $L_{\mathrm{mse}}$ and $L_{\mathrm{per}}$; (c) $L_{\mathrm{per}}$ only. Each example shows a single channel to enhance noise artifact visualization.}
    \label{fig:reconstruction_loss}
\end{figure}

\begin{table}[t]
    \centering
    \begin{tabular}{C{1.5}C{1.0}C{1.0}C{1.6}C{1.0}}
        \toprule
        $L_{\mathrm{mse}}$ : $L_{\mathrm{per}}$ & Metric   & MNIST & CelebA-HQ  & AF-C \\ \midrule
        \multirow{2}{*}{10 : 0} 
            & PSNR($\uparrow$)  & 33.53 & 26.91 & 23.60  \\ 
            & SSIM($\uparrow$)  & \! \! \! \! \! \! 0.76  & \! \! \! \! \! \! 0.78  & \! \! \! \! \! \! 0.60   \\ \midrule
        \multirow{2}{*}{9 : 1} 
            & PSNR($\uparrow$)  & 31.86 & 24.95 & 22.70  \\ 
            & SSIM($\uparrow$)  & \! \! \! \! \! \! 0.71  & \! \! \! \! \! \! 0.73  & \! \! \! \! \! \! 0.61   \\ \midrule
        \multirow{2}{*}{0 : 10} 
            & PSNR($\uparrow$)  & 23.09 & 22.90 & 12.67  \\ 
            & SSIM($\uparrow$)  & \! \! \! \! \! \! 0.41  & \! \! \! \! \! \! 0.54  & \! \! \! \! \! \! 0.41   \\ \bottomrule
    \end{tabular}%
    \caption{The average PSNR and SSIM scores for P-Hologen, trained with various $L_{\mathrm{mse}}$ : $L_{\mathrm{per}}$ ratios.}
    \label{tab:recon_ratio_validation}
\end{table}

\subsection{Reconstruction loss}
\label{subsec:recon_loss}

In image synthesis, generative models often employ either MSE loss $L_{\mathrm{mse}}$ or perceptual loss $L_{\mathrm{per}}$ to enhance reconstruction quality.
$L_{\mathrm{mse}}$ aims at minimizing pixel-level discrepancies between the generated and target images, thereby achieving high fidelity in pixel-value correspondence. 
Conversely, $L_{\mathrm{per}}$~\cite{johnson2016perceptual}, aims to enhance perceptual similarity between images, aligning more congruently with human visual processing mechanisms.

\begin{figure}[t]
    \centering
    \includegraphics[width=\columnwidth]{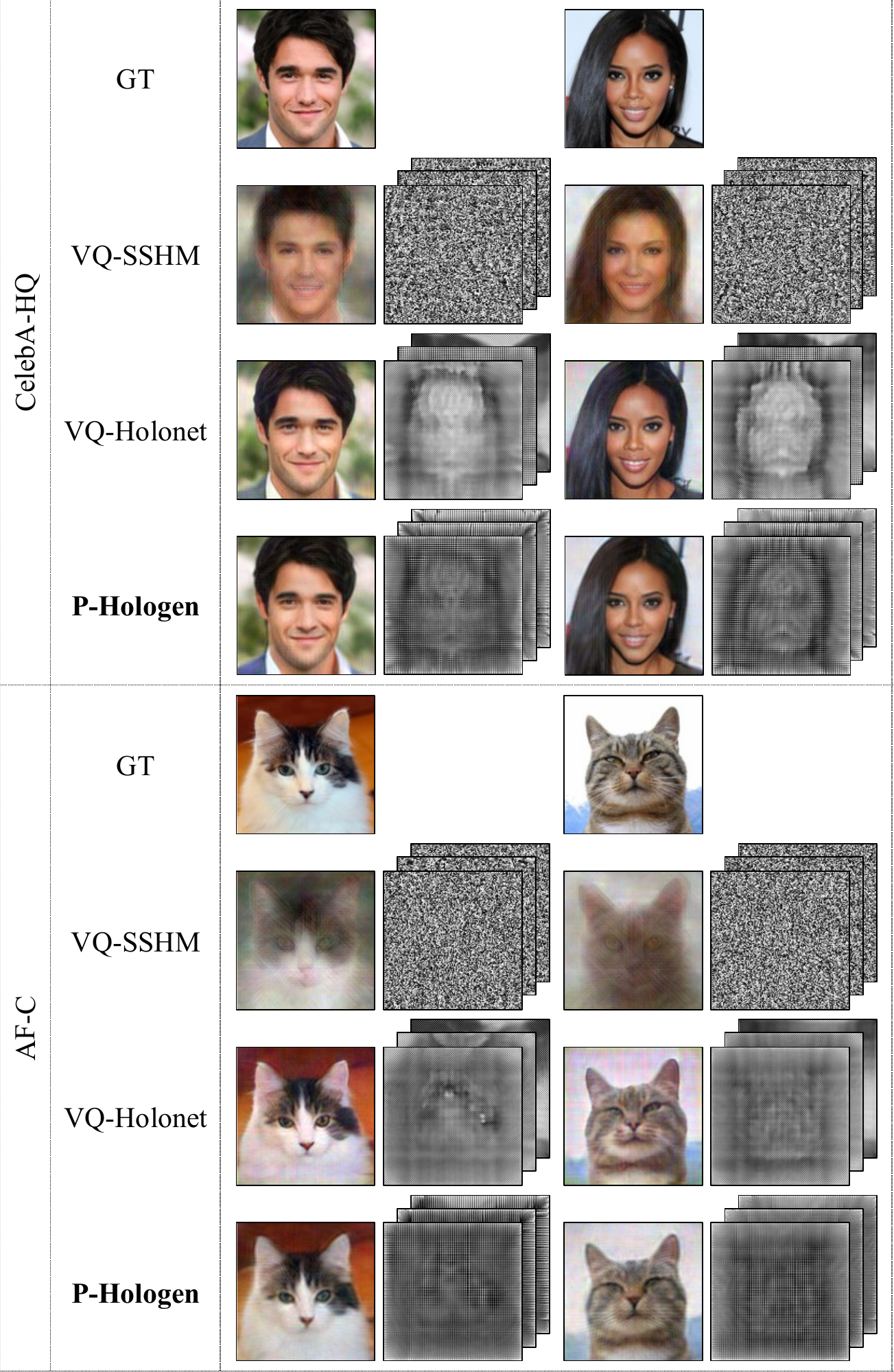}
    \caption{POHs and their reconstructions of each model using GT images from CelebA-HQ and AF-C datasets.}
    \label{fig:recons}
\end{figure}

\begin{table}[t]
    \centering
    \begin{tabular}{C{1.5}C{1.0}C{1.0}C{1.6}C{1.0}}
        \toprule
        Architecture           & Metric     & MNIST    & CelebA-HQ    & AF-C \\ \midrule
        \multirow{2}{*}{VQ-SSHM} 
            & PSNR($\uparrow$)   & 23.34    & 17.25 & 15.14 \\ 
            & SSIM($\uparrow$)   & \! \! \! \! \! \! 0.66     & \! \! \! \! \! \! 0.47  & \! \! \! \! \! \! 0.31 \\ \midrule
        \multirow{2}{*}{VQ-Holonet} 
            & PSNR($\uparrow$)   & 30.54    & 23.15 & 21.21 \\ 
            & SSIM($\uparrow$)   & \! \! \! \! \! \! 0.68     & \! \! \! \! \! \! 0.69  & \! \! \! \! \! \! 0.58 \\ \midrule
        \multirow{2}{*}{\textbf{P-Hologen}} 
            & PSNR($\uparrow$)   & \textbf{31.86} & \textbf{24.95} & \textbf{22.70} \\ 
            & SSIM($\uparrow$)   & \! \! \! \! \! \! \textbf{0.71}  & \! \! \! \! \! \! \textbf{0.73}  & \! \! \! \! \! \! \textbf{0.61} \\ \bottomrule
    \end{tabular}%
    \caption{POH reconstruction accuracy comparison of models trained on MNIST, CelebA-HQ, and AF-C datasets.}
    \label{tab:quantitative}
\end{table}

Through the experiment, we observed an additional benefit of employing $L_{\mathrm{per}}$ for POH generation: it helps suppress noise in the reconstructed images, a critical issue in this domain~\cite{chang2015speckle, chang2017speckle}. 
When the model was trained solely with $L_{\mathrm{mse}}$, the reconstructed image exhibited noticeable noise, as shown in~\autoref{fig:reconstruction_loss}(a). Conversely, training the model exclusively with $L_{\mathrm{per}}$ resulted in smoother images with reduced noise, as in~\autoref{fig:reconstruction_loss}(c).

However, relying solely on $L_{\mathrm{per}}$ led to a decline in the average PSNR and SSIM scores for reconstruction, as shown in~\autoref{tab:recon_ratio_validation}.
To mitigate this issue, we combined $L_{\mathrm{mse}}$ with $L_{\mathrm{per}}$ in a 9:1 ratio, as demonstrated in~\autoref{fig:reconstruction_loss}(b). This combination was empirically determined to balance the PSNR and SSIM metrics with perceptual quality, thereby optimizing the overall reconstruction performance. 

\subsection{Comparative analysis}
\label{subsec:comparative}

This section presents our comparative analysis with two generative methods for novel POH generation: VQ-SSHM, which employs the SSHMs introduced by Liu \textit{et al.}~\cite{liu2021deep}, and VQ-Holonet, an integrated method combining an image-domain generative model with an image-to-hologram conversion model.
To ensure a fair comparison, VQ-SSHM is based on our VQ-VAE architecture, which, as observed in our internal tests, surpasses the performance of the original VAE implementation.
VQ-Holonet utilizes our VQ-VAE for the image-domain generative model and Holonet \cite{peng2020neural} for the image-to-hologram conversion.
VQ-VAE used for both VQ-SSHM and VQ-Holonet employs the same reconstruction loss as P-Hologen. 
The Holonet in VQ-Holonet is trained according to the methodology outlined by Peng \textit{et al.}~\cite{peng2020neural}, with modifications to exclude the camera-in-the-loop component due to practical constraints.

The comparison was conducted in terms of reconstruction accuracy, novel POH generation capability, and computational efficiency.
The datasets of CelebA-HQ and AF-C were utilized for training the models, chosen for their rich and diverse features.
The MNIST dataset was included exclusively for evaluating reconstruction accuracy to provide a comprehensive comparison. It was excluded from generation capability evaluation due to the unconditional nature of our model and the large class variability in the MNIST dataset, which results in fluctuations in FID scores.

\subsubsection{Reconstruction accuracy}
\label{subsubsec:acc_eval}

In this section, we evaluate the accuracy of the reconstructed POHs using PSNR and SSIM. 
These metrics are computed and averaged across all images in the validation set. 
For illustrative purposes, we selected two representative ground truth (GT) images from each dataset, as shown in \autoref{fig:recons}.

While the reconstructions produced by VQ-SSHM capture essential features of the ground truth images and generate plausible outputs, they fail to capture high-level details, resulting in blurred images. 
Conversely, the reconstructions from VQ-Holonet and P-Hologen demonstrate superior proficiency in generating clearer results that effectively restore high-level details. 

We hypothesize that while SSHMs were sufficient for the model to learn target features from simple datasets, such as those used by Liu \textit{et al.}~\cite{liu2021deep} (50 images), the learning process becomes significantly more challenging with higher complexity datasets, such as ours (over 5000 images).
In contrast, the competitive results of P-Hologen compared to VQ-Holonet indicate that our image-based loss design enables generative models to learn the intricate features necessary for high-quality POH generation, even with complex datasets.

\begin{figure*}[t]
    \centering
    \includegraphics[width=\textwidth]{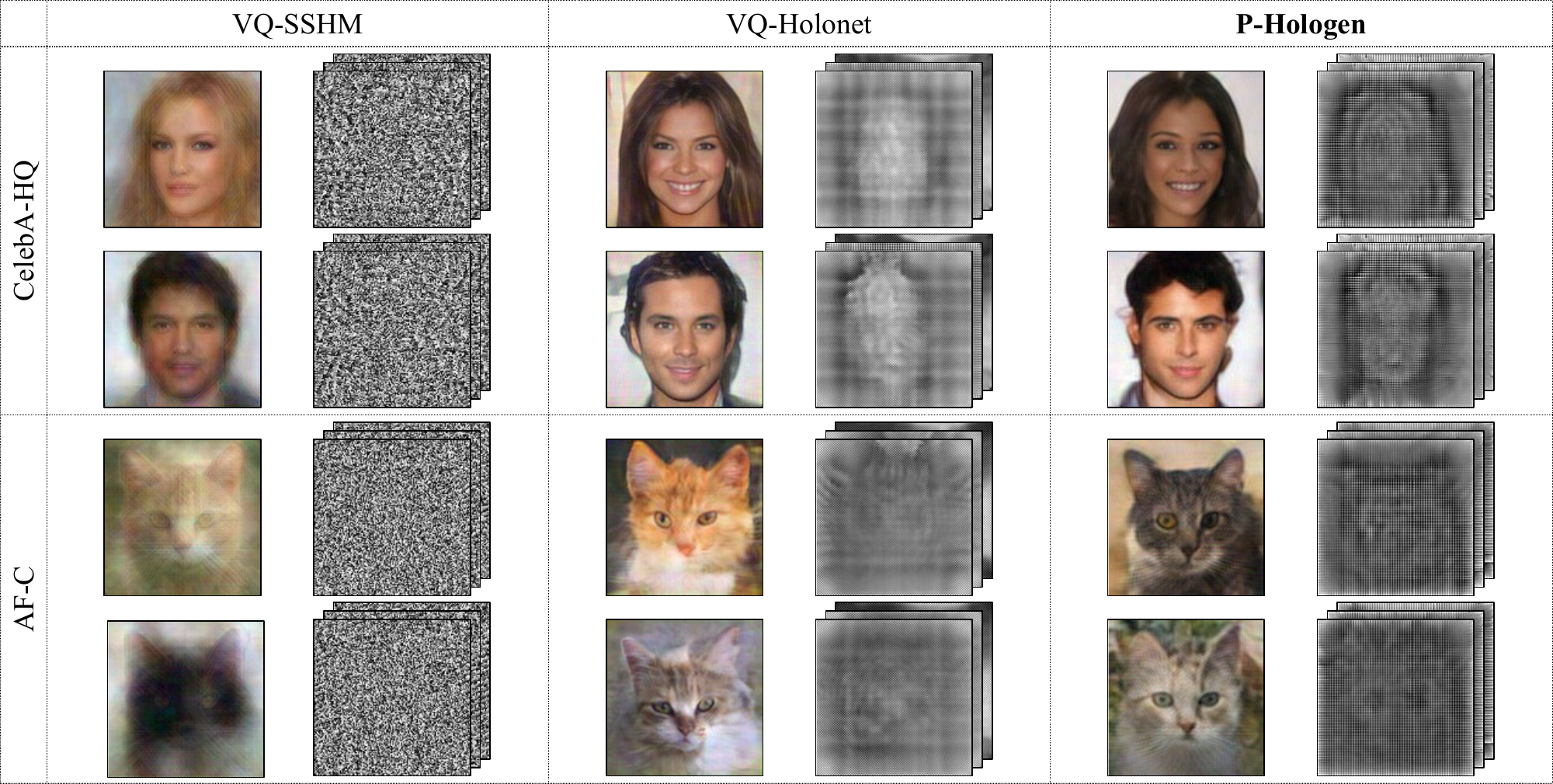}
    \caption{Representative novel POHs and their reconstructions of models trained on CelebA-HQ and AF-C datasets.}
    \label{fig:samples}
\end{figure*}

\begin{table}[t]
    \centering
    \begin{tabular}{C{2.2}C{1.4}C{1.6}C{1.4}}
        \toprule
        Architecture     & Metric            & CelebA-HQ    & AF-C \\ \midrule
        VQ-SSHM          & \multirow{3}{*}{FID($\downarrow$)}  & 169.36  & 161.75    \\ 
        VQ-Holonet       &                                     & 139.84  & 151.02    \\ 
        \textbf{P-Hologen} &                                  & \textbf{126.21}  &  \textbf{130.30}    \\ \bottomrule
    \end{tabular}%
    \caption{FID scores for reconstructions of novel POHs from models trained on CelebA-HQ and AF-C.}
    \label{tab:fids}
\end{table}

The quantitative evaluation results presented in~\autoref{tab:quantitative} support this assertion, as both PSNR and SSIM scores for VQ-Holonet and P-Hologen are significantly higher compared to those of the VQ-SSHM, with our method showing a slight advantage.

\subsubsection{Novel POH generation capability}
\label{subsubsec:gen_eval}
The core functionality of a generative model is to generate new, unseen data instances that closely resemble those in the training dataset.
We quantitatively assessed the effectiveness of each approach using Fréchet Inception Distance (FID) scores, which were obtained by comparing 100 generated samples to a randomly selected set of 100 images from the validation set. 
The results of this analysis are depicted in~\autoref{fig:samples}, with the corresponding FID scores detailed in \autoref{tab:fids}.

Similar to the reconstruction task, VQ-SSHM often generates novel POHs whose reconstructions lack high-level detail and appear blurred. 
In contrast, the novel POHs generated from VQ-Holonet and P-Hologen result in reconstructions that are clear and rich in high-level details.

Consistent with qualitative comparisons, VQ-Holonet and P-Hologen exhibit superior performance in terms of FID scores, as indicated in \autoref{tab:fids}.

To further validate the uniqueness of the samples generated by P-Hologen, we present images from the training set that exhibit the highest similarity to the generated samples, commonly referred to as nearest neighbors.
The similarity between the generated samples and the training set images was computed leveraging the LPIPS metric.
If our model had merely memorized the inputs from the training set rather than learning their features, the generated samples would be identical to the images in the training set.

\begin{figure}[t]
    \centering
    \includegraphics[width=\columnwidth]{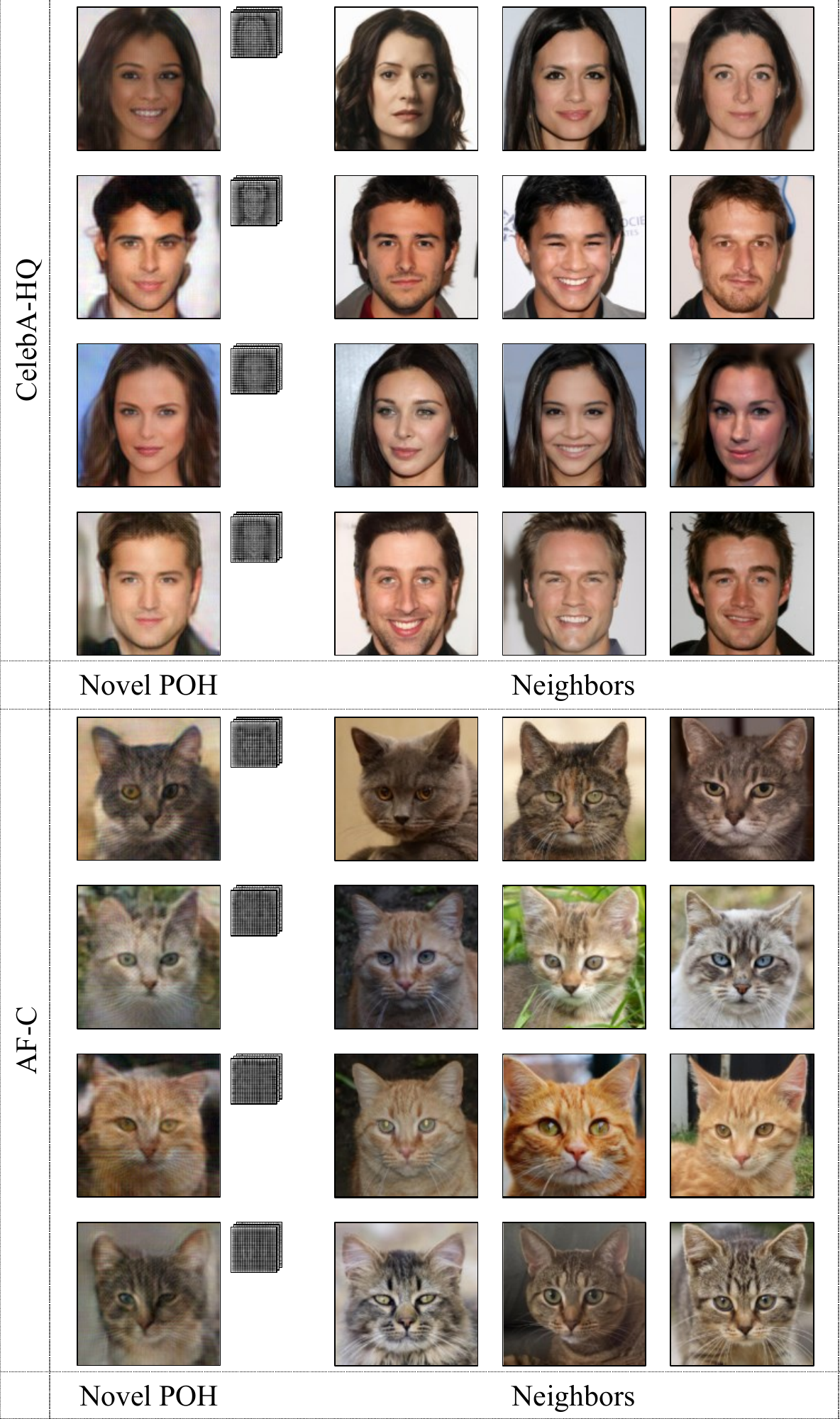}
    \caption{Nearest neighbors of the generated novel POHs of P-Hologen, computed using the similarity metric of LPIPS.}
    \label{fig:neighbors}
\end{figure}

As shown in~\autoref{fig:neighbors}, the generated samples exhibit distinct features from their top three nearest neighbors.
This distinction confirms our model’s ability to generalize and produce novel data instances that are similar but not identical to those in the training set.

\subsubsection{Computational efficiency}
\label{subsubsec:comp_eval}
In this section, we examine the computational requirements of each approach, focusing on the number of network parameters and the sampling time necessary for generating novel POHs at a resolution of $128\times128$.
Considering the variability in sampling time based on the implemented sampler, we evaluate sampling time in two ways: decoding time, which is the time required for the decoder architecture to process the sampled latent vector, and inference time, which is the sum of the decoding time and the time needed for the sampler to generate a sample latent vector.

\begin{table}[t]
    \centering
    \begin{tabular}{C{2.2}C{1.4}C{1.6}C{1.4}}
        \toprule
        Architecture           & Parameters            & Decoding    & Inference \\ \midrule
        VQ-SSHM             & 79.6M                 & 10ms             & 3.810s \\ 
        VQ-Holonet          & 82.5M                 & 13ms             & 3.813s \\ 
        \textbf{P-Hologen}  & \textbf{79.6M}        & \! \! \! \! \! \! \!  \textbf{9ms}     & \textbf{3.809s} \\ \bottomrule
    \end{tabular}
    \caption{Comparison of computational efficiency by network parameters, decoding time, and inference time.}
    \label{tab:computational_load}
\end{table}

As shown in~\autoref{tab:computational_load}, P-Hologen has approximately 3.5\% fewer parameters and a 30\% shorter decoding time compared to VQ-Holonet.
Given that P-Hologen presents POHs of comparable quality, this reduction in computational demand indicates its promising efficiency.
Although this gap may seem negligible in the context of total inference time, selecting a more advanced sampler model can further reduce inference time, accentuating the disparity.
Moreover, as P-Hologen is an end-to-end model, it requires only a single training process, unlike VQ-Holonet, which necessitates training two separate models.

While VQ-SSHM has similar computational demands to P-Hologen, the quality of the generated POHs is not as high, as shown in \autoref{subsubsec:acc_eval} and \autoref{subsubsec:gen_eval}. 
Additionally, the training dataset must be preprocessed into SSHMs before training, which can be cumbersome for users.

\begin{figure}[t] 
    \centering
    \includegraphics[width=\columnwidth]{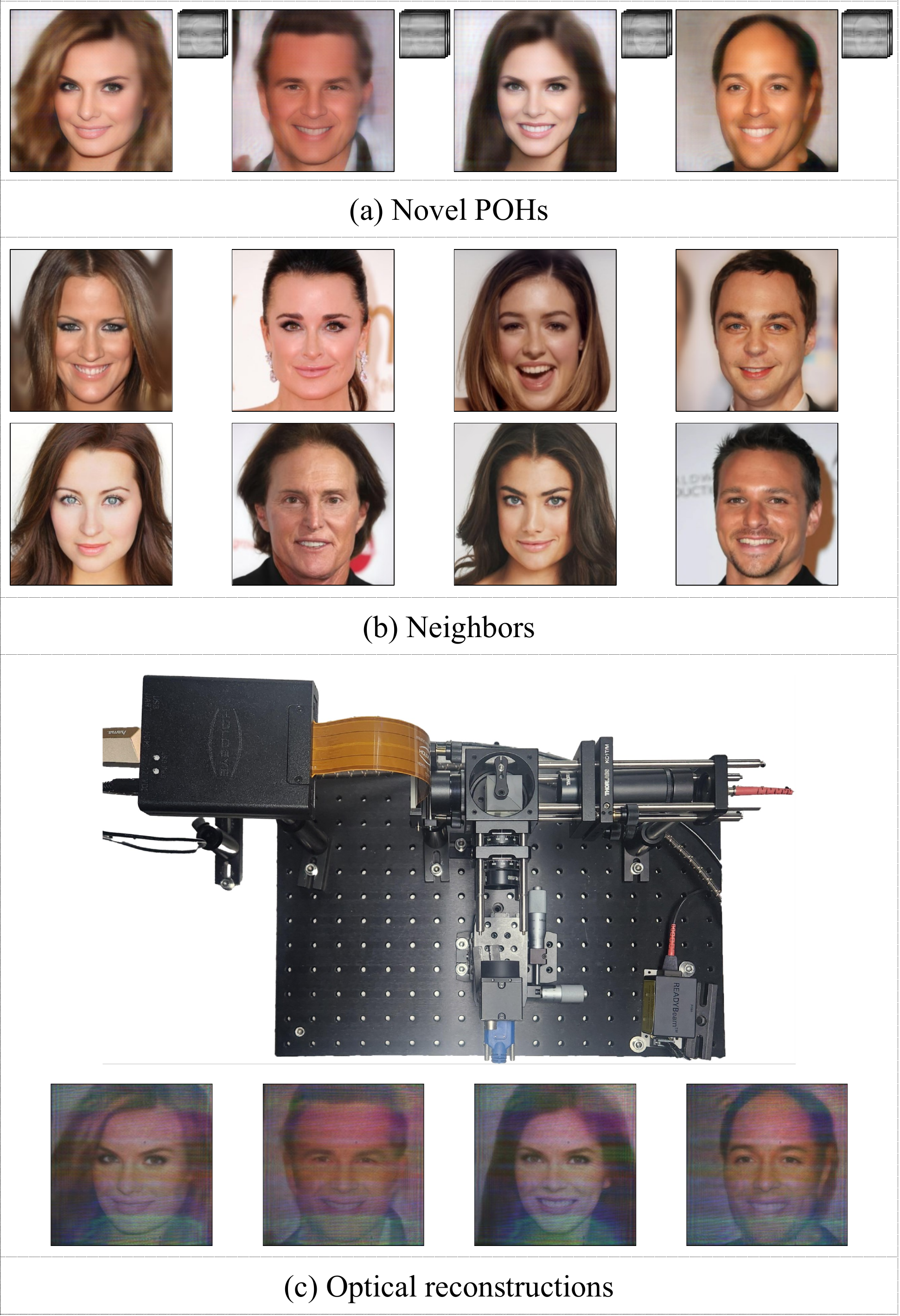}
    \caption{(a) High-resolution novel POHs generated by P-Hologen trained on CelebA-HQ dataset. (b) top two nearest neighbors. (c) holography setup and the optical reconstruction results.}
    \label{fig:optical_recons}
\end{figure}

\subsection{Optical reconstructions}
\label{subsec:optical_recons}
Our holography setup uses FISBA RGBeam fiber-coupled module with laser diodes of three wavelengths. 
The laser is collimated with Thorlabs AC254-100-A-ML lens of 100mm focal length and KC1T/M is used to control the orientation of the lens. 
Collimation is validated by a Shear Interferometer, Thorlabs SI254. 
We use Holoeye Leto-3 Phase Only Spatial Light Modulator which has reflective LCOS display of $1920\times1080$ resolution with a pixel pitch of 6.4 $\mu m$ and 8 bits precision, with the propagation distance set to 122.3 \textit{mm}.
Orientation of the SLM is adjusted by Thorlabs PY005/M.
Images are captured via FLIR BFLY-U3-23S6C-C USB 3.0 Blackfly camera with Sony IMX249 color sensor of $1920\times1200$ resolution.

We trained our model on the CelebA-HQ dataset at a resolution of $1024\times1024$, with other configurations following our device specifications.
The training was conducted for 100 epochs using two NVIDIA RTX A6000 GPUs.
Aligning with our POH resolution, the central $1024\times1024$ region from the captured images are cropped for better visualization.
To present the performance of our work in high resolution and optical reconstruction, the generated POHs, their top two nearest neighbors, and their optical reconstruction results are listed in~\autoref{fig:optical_recons}.

\section{Limitations and future work}
\label{sec:limit_future}

Addressing the limitations and future directions of our research involves four key aspects.

First, our model's FID scores are relatively high compared to recent image generation models due to the error sensitivity and the presence of noise, as discussed in \autoref{subsec:recon_loss}. 
We anticipate that a more sophisticated learning strategy will improve FID scores. For example, utilizing the multi-level latent space of the VQ-VAE2~\cite{razavi2019generating} architecture to separate global and local features may reduce noise, though it would increase computational costs due to learning and sampling from multiple latent spaces.

Second, our model generates POHs optimized for a fixed propagation distance $p_{d}$, limiting effectiveness for viewers at varying distances. Developing methods to accommodate variable propagation distances will enhance the versatility of our models. Based on our contribution to constructing phase latent space, we will investigate POH generative models across a range of distances to improve real-world applicability.

Third, while adopting perceptual loss reduces noise in the reconstructed image, it also decreases the ability to capture bright pixels, as shown in \autoref{fig:reconstruction_loss}. This occurs because perceptual loss, focused on high-level features from the pretrained VGG network, does not enforce strict pixel-wise accuracy, leading to slight intensity deviations compared to MSE loss. Utilizing the frequency loss introduced in \cite{tseng2024neural} may address this by pushing reconstruction noise outside human-perceivable frequency bands, preserving the critical frequencies needed for high-quality image reproduction.

Lastly, the generated POHs lack accommodation cues, which are crucial for depth perception and realism in holographic displays. 
We plan to incorporate these cues in our future generative models. 

\section{Conclusion}
\label{sec:conclusion}

Recent image generation methods leveraging generative models have unlocked a plethora of creative possibilities, ranging from seamless image interpolation to intricate text-driven modifications. The cornerstone of these advancements lies in the effective utilization of latent spaces which distilled the core features of target data.

Despite these advancements in image generation, the field of holography, particularly a generative model for hologram generation, has yet to fully embrace the potential of latent spaces. 
This omission has constrained the holography domain from accessing the advanced capabilities inherent in generative models, limiting the scope of innovation and creativity in hologram generation.

To bridge this gap, we present P-Hologen, a novel framework designed to integrate the generative models into the domain of hologram generation. 
As a pioneering effort, P-Hologen marks a significant stride towards enriching the holography field with the advanced functionalities of generative models. 
This suggests the potential for advanced features like style transfer and text-driven editing to enrich holographic content, as well as benefit the field of computer graphics by providing new tools for generating and manipulating 3D content, leading to more realistic VR and AR experiences and more sophisticated visual effects in digital media.

\section*{Acknowledgments}
\label{sec:ack} 
This work was supported by Samsung Research Funding \& Incubation Center of Samsung Electronics under Project Number SRFC-IT2201-03 (80\%), ICT Creative Consilience Program through the Institute of Information \& Communications Technology Planning \& Evaluation(IITP) grant funded by the Korea government(MSIT) (RS-2020-II201819, 10\%), and  National Research Foundation of Korea (NRF) grant funded by the Korea government (MSIT) (No. RS-2023-00211658, 10\%)

\bibliographystyle{eg-alpha-doi} 
\bibliography{egbibsample}       

\newcommand{\etalchar}[1]{$^{#1}$}
\begin{thebibliography}{\uppercase{MYMVL14}}

\bibitem[BCC{\etalchar{*}}24]{ban2024nhvc}
\textsc{Ban H., Choi S., Cha J.~Y., Kim Y., Kim H.~Y.}:
\newblock Nhvc: Neural holographic video compression with scalable architecture.
\newblock In \emph{2024 IEEE Conference Virtual Reality and 3D User Interfaces (VR)} (2024), IEEE, pp.~969--978.

\bibitem[Ben94]{bengtsson1994kinoform}
\textsc{Bengtsson J.}:
\newblock Kinoform design with an optimal-rotation-angle method.
\newblock \emph{Applied optics 33}, 29 (1994), 6879--6884.

\bibitem[Bla21]{blanche2021holography}
\textsc{Blanche P.-A.}:
\newblock Holography, and the future of 3d display.
\newblock \emph{Light: Advanced Manufacturing 2}, 4 (2021), 446--459.

\bibitem[CGP{\etalchar{*}}21]{choi2021neural}
\textsc{Choi S., Gopakumar M., Peng Y., Kim J., Wetzstein G.}:
\newblock Neural 3d holography: Learning accurate wave propagation models for 3d holographic virtual and augmented reality displays.
\newblock \emph{ACM Transactions on Graphics (TOG) 40}, 6 (2021), 1--12.

\bibitem[CMRA18]{chen2018pixelsnail}
\textsc{Chen X., Mishra N., Rohaninejad M., Abbeel P.}:
\newblock Pixelsnail: An improved autoregressive generative model.
\newblock In \emph{International Conference on Machine Learning} (2018), PMLR, pp.~864--872.

\bibitem[CPK{\etalchar{*}}19]{chakravarthula2019wirtinger}
\textsc{Chakravarthula P., Peng Y., Kollin J., Fuchs H., Heide F.}:
\newblock Wirtinger holography for near-eye displays.
\newblock \emph{ACM Transactions on Graphics (TOG) 38}, 6 (2019), 1--13.

\bibitem[CQW{\etalchar{*}}17]{chang2017speckle}
\textsc{Chang C., Qi Y., Wu J., Xia J., Nie S.}:
\newblock Speckle reduced lensless holographic projection from phase-only computer-generated hologram.
\newblock \emph{Optics Express 25}, 6 (2017), 6568--6580.

\bibitem[CTS{\etalchar{*}}20]{chakravarthula2020learned}
\textsc{Chakravarthula P., Tseng E., Srivastava T., Fuchs H., Heide F.}:
\newblock Learned hardware-in-the-loop phase retrieval for holographic near-eye displays.
\newblock \emph{ACM Transactions on Graphics (TOG) 39}, 6 (2020), 1--18.

\bibitem[CUYH20]{choi2020stargan}
\textsc{Choi Y., Uh Y., Yoo J., Ha J.-W.}:
\newblock Stargan v2: Diverse image synthesis for multiple domains.
\newblock In \emph{Proceedings of the IEEE/CVF conference on computer vision and pattern recognition} (2020), pp.~8188--8197.

\bibitem[CXY{\etalchar{*}}15]{chang2015speckle}
\textsc{Chang C., Xia J., Yang L., Lei W., Yang Z., Chen J.}:
\newblock Speckle-suppressed phase-only holographic three-dimensional display based on double-constraint gerchberg--saxton algorithm.
\newblock \emph{Applied optics 54}, 23 (2015), 6994--7001.

\bibitem[DBS96]{dresel1996design}
\textsc{Dresel T., Beyerlein M., Schwider J.}:
\newblock Design of computer-generated beam-shaping holograms by iterative finite-element mesh adaption.
\newblock \emph{Applied Optics 35}, 35 (1996), 6865--6874.

\bibitem[ERO21]{esser2021taming}
\textsc{Esser P., Rombach R., Ommer B.}:
\newblock Taming transformers for high-resolution image synthesis.
\newblock In \emph{Proceedings of the IEEE/CVF conference on computer vision and pattern recognition} (2021), pp.~12873--12883.

\bibitem[Ger72]{gerchberg1972practical}
\textsc{Gerchberg R.~W.}:
\newblock A practical algorithm for the determination of plane from image and diffraction pictures.
\newblock \emph{Optik 35}, 2 (1972), 237--246.

\bibitem[GLZ{\etalchar{*}}23]{gao2023implicit}
\textsc{Gao S., Liu X., Zeng B., Xu S., Li Y., Luo X., Liu J., Zhen X., Zhang B.}:
\newblock Implicit diffusion models for continuous super-resolution.
\newblock In \emph{Proceedings of the IEEE/CVF conference on computer vision and pattern recognition} (2023), pp.~10021--10030.

\bibitem[Goo05]{goodman2005introduction}
\textsc{Goodman J.~W.}:
\newblock \emph{Introduction to Fourier optics}.
\newblock Roberts and Company publishers, 2005.

\bibitem[HECA{\etalchar{*}}20]{hossein2020deepcgh}
\textsc{Hossein~Eybposh M., Caira N.~W., Atisa M., Chakravarthula P., P{\'e}gard N.~C.}:
\newblock Deepcgh: 3d computer-generated holography using deep learning.
\newblock \emph{Optics Express 28}, 18 (2020), 26636--26650.

\bibitem[JAFF16]{johnson2016perceptual}
\textsc{Johnson J., Alahi A., Fei-Fei L.}:
\newblock Perceptual losses for real-time style transfer and super-resolution.
\newblock In \emph{Computer Vision--ECCV 2016: 14th European Conference, Amsterdam, The Netherlands, October 11-14, 2016, Proceedings, Part II 14} (2016), Springer, pp.~694--711.

\bibitem[JJC{\etalchar{*}}18]{jiao2018compression}
\textsc{Jiao S., Jin Z., Chang C., Zhou C., Zou W., Li X.}:
\newblock Compression of phase-only holograms with jpeg standard and deep learning.
\newblock \emph{Applied Sciences 8}, 8 (2018), 1258.

\bibitem[JK08]{jabbour2008vectorial}
\textsc{Jabbour T.~G., Kuebler S.~M.}:
\newblock Vectorial beam shaping.
\newblock \emph{Optics Express 16}, 10 (2008), 7203--7213.

\bibitem[KALL17]{karras2017progressive}
\textsc{Karras T., Aila T., Laine S., Lehtinen J.}:
\newblock Progressive growing of gans for improved quality, stability, and variation.
\newblock \emph{arXiv preprint arXiv:1710.10196} (2017).

\bibitem[KPK{\etalchar{*}}21]{kang2021deep}
\textsc{Kang J.-W., Park B.-S., Kim J.-K., Kim D.-W., Seo Y.-H.}:
\newblock Deep-learning-based hologram generation using a generative model.
\newblock \emph{Applied Optics 60}, 24 (2021), 7391--7399.

\bibitem[KUA22]{kavakli2022learned}
\textsc{Kavakl{\i} K., Urey H., Ak{\c{s}}it K.}:
\newblock Learned holographic light transport.
\newblock \emph{Applied Optics 61}, 5 (2022), B50--B55.

\bibitem[LC21]{liu2021deep}
\textsc{Liu S.-C., Chu D.}:
\newblock Deep learning for hologram generation.
\newblock \emph{Optics Express 29}, 17 (2021), 27373--27395.

\bibitem[LCB{\etalchar{*}}10]{lecun2010mnist}
\textsc{LeCun Y., Cortes C., Burges C., et~al.}:
\newblock Mnist handwritten digit database, 2010.

\bibitem[LP67]{lohmann1967binary}
\textsc{Lohmann A.~W., Paris D.}:
\newblock Binary fraunhofer holograms, generated by computer.
\newblock \emph{Applied optics 6}, 10 (1967), 1739--1748.

\bibitem[MS09]{matsushima2009band}
\textsc{Matsushima K., Shimobaba T.}:
\newblock Band-limited angular spectrum method for numerical simulation of free-space propagation in far and near fields.
\newblock \emph{Optics express 17}, 22 (2009), 19662--19673.

\bibitem[MYMVL14]{mendoza2014encoding}
\textsc{Mendoza-Yero O., M{\'\i}nguez-Vega G., Lancis J.}:
\newblock Encoding complex fields by using a phase-only optical element.
\newblock \emph{Optics letters 39}, 7 (2014), 1740--1743.

\bibitem[PCPW20]{peng2020neural}
\textsc{Peng Y., Choi S., Padmanaban N., Wetzstein G.}:
\newblock Neural holography with camera-in-the-loop training.
\newblock \emph{ACM Transactions on Graphics (TOG) 39}, 6 (2020), 1--14.

\bibitem[PDSH17]{peng2017mix}
\textsc{Peng Y., Dun X., Sun Q., Heidrich W.}:
\newblock Mix-and-match holography.
\newblock \emph{ACM Trans. Graph. 36}, 6 (2017), 191--1.

\bibitem[PWZ{\etalchar{*}}17]{pang2017non}
\textsc{Pang H., Wang J., Zhang M., Cao A., Shi L., Deng Q.}:
\newblock Non-iterative phase-only fourier hologram generation with high image quality.
\newblock \emph{Optics Express 25}, 13 (2017), 14323--14333.

\bibitem[RBM{\etalchar{*}}19]{rad2019srobb}
\textsc{Rad M.~S., Bozorgtabar B., Marti U.-V., Basler M., Ekenel H.~K., Thiran J.-P.}:
\newblock Srobb: Targeted perceptual loss for single image super-resolution.
\newblock In \emph{Proceedings of the IEEE/CVF international conference on computer vision} (2019), pp.~2710--2719.

\bibitem[RVdOV19]{razavi2019generating}
\textsc{Razavi A., Van~den Oord A., Vinyals O.}:
\newblock Generating diverse high-fidelity images with vq-vae-2.
\newblock \emph{Advances in neural information processing systems 32} (2019).

\bibitem[SEC{\etalchar{*}}15]{shechtman2015phase}
\textsc{Shechtman Y., Eldar Y.~C., Cohen O., Chapman H.~N., Miao J., Segev M.}:
\newblock Phase retrieval with application to optical imaging: a contemporary overview.
\newblock \emph{IEEE signal processing magazine 32}, 3 (2015), 87--109.

\bibitem[SGZ23]{singh2023high}
\textsc{Singh J., Gould S., Zheng L.}:
\newblock High-fidelity guided image synthesis with latent diffusion models.
\newblock In \emph{2023 IEEE/CVF Conference on Computer Vision and Pattern Recognition (CVPR)} (2023), IEEE, pp.~5997--6006.

\bibitem[SKI15]{shimobaba2015review}
\textsc{Shimobaba T., Kakue T., Ito T.}:
\newblock Review of fast algorithms and hardware implementations on computer holography.
\newblock \emph{IEEE Transactions on Industrial Informatics 12}, 4 (2015), 1611--1622.

\bibitem[SLK{\etalchar{*}}21]{shi2021towards}
\textsc{Shi L., Li B., Kim C., Kellnhofer P., Matusik W.}:
\newblock Towards real-time photorealistic 3d holography with deep neural networks.
\newblock \emph{Nature 591}, 7849 (2021), 234--239.

\bibitem[SLM22]{shi2022end}
\textsc{Shi L., Li B., Matusik W.}:
\newblock End-to-end learning of 3d phase-only holograms for holographic display.
\newblock \emph{Light: Science \& Applications 11}, 1 (2022), 247.

\bibitem[TCP14]{tsang2014generation}
\textsc{Tsang P., Chow Y.-T., Poon T.-C.}:
\newblock Generation of phase-only fresnel hologram based on down-sampling.
\newblock \emph{Optics express 22}, 21 (2014), 25208--25214.

\bibitem[TGBD23]{tumanyan2023plug}
\textsc{Tumanyan N., Geyer M., Bagon S., Dekel T.}:
\newblock Plug-and-play diffusion features for text-driven image-to-image translation.
\newblock In \emph{Proceedings of the IEEE/CVF Conference on Computer Vision and Pattern Recognition} (2023), pp.~1921--1930.

\bibitem[TKB{\etalchar{*}}24]{tseng2024neural}
\textsc{Tseng E., Kuo G., Baek S.-H., Matsuda N., Maimone A., Schiffers F., Chakravarthula P., Fu Q., Heidrich W., Lanman D., et~al.}:
\newblock Neural {\'e}tendue expander for ultra-wide-angle high-fidelity holographic display.
\newblock \emph{Nature communications 15}, 1 (2024), 2907.

\bibitem[TLLP21]{tsang2021optimal}
\textsc{Tsang P., Liu J.-P., Lam H., Poon T.-C.}:
\newblock Optimal sampled phase-only hologram (ospoh).
\newblock \emph{Optics Express 29}, 16 (2021), 25488--25498.

\bibitem[TP13]{tsang2013novel}
\textsc{Tsang P. W.~M., Poon T.-C.}:
\newblock Novel method for converting digital fresnel hologram to phase-only hologram based on bidirectional error diffusion.
\newblock \emph{Optics express 21}, 20 (2013), 23680--23686.

\bibitem[VdMH08]{van2008visualizing}
\textsc{Van~der Maaten L., Hinton G.}:
\newblock Visualizing data using t-sne.
\newblock \emph{Journal of machine learning research 9}, 11 (2008).

\bibitem[VDOV{\etalchar{*}}17]{van2017neural}
\textsc{Van Den~Oord A., Vinyals O., et~al.}:
\newblock Neural discrete representation learning.
\newblock \emph{Advances in neural information processing systems 30} (2017).

\bibitem[WBW{\etalchar{*}}20]{wang2020phase}
\textsc{Wang F., Bian Y., Wang H., Lyu M., Pedrini G., Osten W., Barbastathis G., Situ G.}:
\newblock Phase imaging with an untrained neural network.
\newblock \emph{Light: Science \& Applications 9}, 1 (2020), 77.

\bibitem[WLSC21]{wu2021high}
\textsc{Wu J., Liu K., Sui X., Cao L.}:
\newblock High-speed computer-generated holography using an autoencoder-based deep neural network.
\newblock \emph{Optics Letters 46}, 12 (2021), 2908--2911.

\bibitem[YIY95]{yoshikawa1995quantized}
\textsc{Yoshikawa N., Itoh M., Yatagai T.}:
\newblock Quantized phase optimization of two-dimensional fourier kinoforms by a genetic algorithm.
\newblock \emph{Optics Letters 20}, 7 (1995), 752--754.

\bibitem[YSY{\etalchar{*}}22]{yang2022diffraction}
\textsc{Yang D., Seo W., Yu H., Kim S.~I., Shin B., Lee C.-K., Moon S., An J., Hong J.-Y., Sung G., et~al.}:
\newblock Diffraction-engineered holography: Beyond the depth representation limit of holographic displays.
\newblock \emph{Nature Communications 13}, 1 (2022), 6012.

\bibitem[ZNV{\etalchar{*}}21]{zhang2021phasegan}
\textsc{Zhang Y., Noack M.~A., Vagovic P., Fezzaa K., Garcia-Moreno F., Ritschel T., Villanueva-Perez P.}:
\newblock Phasegan: a deep-learning phase-retrieval approach for unpaired datasets.
\newblock \emph{Optics express 29}, 13 (2021), 19593--19604.

\bibitem[ZZL21]{zeng2021deep}
\textsc{Zeng T., Zhu Y., Lam E.~Y.}:
\newblock Deep learning for digital holography: a review.
\newblock \emph{Optics Express 29}, 24 (2021), 40572--40593.

\end{thebibliography}



\end{document}